\documentclass[a4paper]{article}
\usepackage{INTERSPEECH2022}

\usepackage{amsmath}
\usepackage{amsfonts}
\usepackage{graphicx}
\usepackage{booktabs}
\usepackage{multirow}
\usepackage{caption}
\usepackage{subcaption}

\usepackage{pifont}
\newcommand{\cmark}{\ding{51}}

\title{Similarity and Content-based Phonetic Self Attention for Speech Recognition}
\name{Kyuhong Shim and Wonyong Sung}
\address{Department of Electrical and Computer Engineering, Seoul National University, Korea}
\email{\{skhu20, wysung\}@snu.ac.kr}

\begin{document}
\maketitle

% ======================================================================================== %
\begin{abstract}
Transformer-based speech recognition models have achieved great success due to the self-attention (SA) mechanism that utilizes every frame in the feature extraction process.
Especially, SA heads in lower layers capture various phonetic characteristics by the query-key dot product, which is designed to compute the pairwise relationship between frames.
In this paper, we propose a variant of SA to extract more representative phonetic features.
The proposed phonetic self-attention (phSA) is composed of two different types of phonetic attention; one is similarity-based and the other is content-based.
In short, similarity-based attention captures the correlation between frames while content-based attention only considers each frame without being affected by other frames.
We identify which parts of the original dot product equation are related to two different attention patterns and improve each part with simple modifications.
Our experiments on phoneme classification and speech recognition show that replacing SA with phSA for lower layers improves the recognition performance without increasing the latency and the parameter size.

\end{abstract}

% ======================================================================================== %
% ======================================================================================== %
\section{Introduction}\label{sec:intro}
% ======================================================================================== %

% (Transformer-based ASR)
End-to-end automatic speech recognition (ASR) has made great progress in line with the advances in deep neural networks (DNNs).
Among various architectures, Transformer~\cite{transformer} models have shown state-of-the-art performance~\cite{transformer-transducer, conformer, pushing, pushing-semi, speechstew} in ASR.
Most Transformer-based ASR models stack the same layer multiple times without considering the difference between layer positions, although the behaviors are very different~\cite{yang20i, usefulness, stochastic}.
If we can identify the role of each layer, we can improve the model architecture by exploiting domain-specific knowledge, resulting in a more heterogeneous composition of layers.
However, because end-to-end DNN performs as a black box, it is difficult to design and apply specific modifications for relevant layers.

% (Phonetic relationship)
Recently, a study suggested that the role of self-attention (SA) in Transformer-based ASR models can be distinguished into two types, phonetic and linguistic localization~\cite{understanding}.
Two roles contribute to speech recognition in a row; the ASR system first extracts phonologically meaningful features by reducing the pronunciation variations and then combines such information into textual features to produce natural output sentences.
These two-stage processes, which correspond to phonetic and linguistic localization, seem to be natural because ASR is a many-to-one problem in that multiple speeches can be transcribed as the same text.
The study discovered that phonetic localization mainly appears in lower layers while linguistic localization happens in upper layers~\cite{understanding}, and their attention patterns are also very different.
The findings imply that we can identify layers of a certain role, and we may boost the performance by improving such layers to perform their role better.

% (We focus on improving phonetic behavior)
Among the two types of roles mentioned above, we focus on improving phonetic localization based on a deeper understanding of the behavior.
Here, we call SA heads that perform phonetic localization a phonetic (attention) head.
From the observation of the attention weights produced by phonetic heads, we can separate two distinct types of attention patterns.
The first type is similarity-based phonetic attention that gives a larger attention weight value on similarly pronounced frames. 
For example, frames corresponding to phoneme class `S' often show large attention weight for frames corresponding to `S', `Z', `SH', and vice versa~\cite{understanding}.
The second type is content-based phonetic attention that attends to certain phonemes regardless of the query.
In other words, a certain attention head may be highly optimized for detecting a specific phoneme class.
We suggest that each phonetic head can be more specialized from the decomposition of similarity-based and content-based attention mechanisms.

% (In this paper)
% In this paper, we propose phonetic self-attention (phSA), a variant of SA that is designed to extract useful phonetic characteristics during phonetic localization.
% The proposed phSA is a simple alternative for SA that is responsible for phonetic localization.
In this paper, we propose phonetic self-attention (phSA), a variant of SA that extracts similarity and content-based phonetic features in phonetic localization.
We modify the query-key dot product term inside the SA mechanism to capture similarity and content separately.
In particular, we improve the dot product by (1) decomposing the two terms to remove shared parameters and (2) inserting trainable non-linearity functions.
We evaluate the proposed phSA using phoneme classification and speech recognition and achieve considerably improved recognition performance on both tasks.
In addition, we empirically show that similarity-based and content-based phonetic attention produce relatively concentrated and distributed attention probabilities, respectively.
% ======================================================================================== %
% ======================================================================================== %
\section{Motivation}\label{sec:motivation}
% ======================================================================================== %

\begin{figure}[t]
    \centering
    \resizebox{1.0\columnwidth}{!}{
    \includegraphics{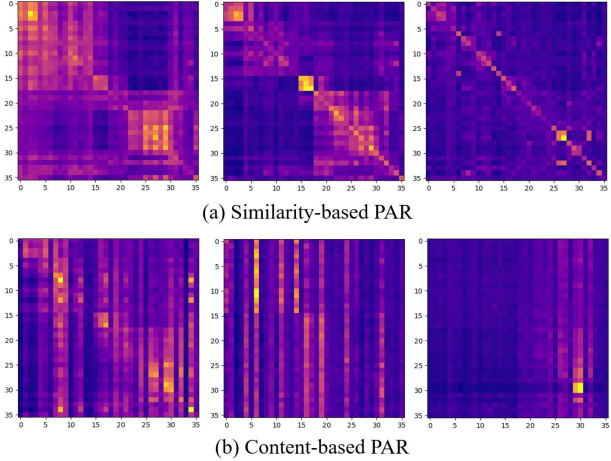}}
    \caption{Visualization of PAR from selected attention heads in the baseline model.
    Two rows show representative examples of similarity-based and content-based phonetic attention, respectively.
    Brighter points indicate higher attention weight.}
    % \vspace{-0.2cm}
    \label{fig:par}
\end{figure}

% ======================================================================================== %
\subsection{Dot Product in Self Attention}\label{ssec:sa}
% ======================================================================================== %

Self-attention (SA) is the key component of Transformer that computes the relationship between every pair of frames.
For a sequence of speech frame features $X=\{x_1, x_2, ... x_T\}$ as an input, SA first projects features into three components, namely query ($Q$), key ($K$), and value ($V$).
SA utilizes multiple attention heads with different parameters to capture diverse relationships in each layer.
Without loss of generality, we explain the behavior of a single attention head below.
$Q$, $K$, and $V$ are linear projections of input as follows:
\begin{equation}
    Q,K,V = XW_{\{Q,K,V\}} + b_{\{Q,K,V\}}
\end{equation}
where $X\in\mathbb{R}^{T\times d_h}$, $W\in\mathbb{R}^{d_h\times d_h}$ and $b\in\mathbb{R}^{1\times d_h}$ are input, weight, and bias, respectively.
$d_h$ is the dimension of each attention head.
The attention map $A$ is then calculated as:
\begin{equation}
    A = \text{Softmax}\left( \frac{Q K^T}{\sqrt{d_h}} \right) \in \mathbb{R}^{T\times T}.
\end{equation}
Each element of the attention map represents how much one frame focuses on the other one, which is, in practice, implemented as a dot product of the query and key.
The dot product equation can be decomposed into four terms:
\begin{equation}
    QK^T = XW_Q W_K^T X^T + XW_Q b_K^T + b_Q W_K^T X^T + b_Q b_K^T.
\end{equation}
The first term ($\in \mathbb{R}^{T\times T}$) calculates the correlation between frames.
The second term ($\in \mathbb{R}^{T\times 1}$) adds offset value per row, while the third term ($\in \mathbb{R}^{1\times T}$) adds offset per column.
The fourth term is a constant.
Because the dot product is followed by the row-wise softmax operation, the second and the fourth terms do not affect the output after softmax.
In other words, the bias of $K$ ($b_K$) can be safely removed, and then the dot-product can be simplified as follows:
\begin{align}
    QK^T &= (XW_Q + b_Q)(XW_K)^T \\
         &= (XW_Q)(XW_K)^T + (XW_K b_Q^T)^T. \label{eq:new_dp}
\end{align}
Please note that the second term of Eq.~\eqref{eq:new_dp} had not been studied much compared to the first term.

% ======================================================================================== %
\subsection{Phonetic Behavior of Self Attention}\label{ssec:phonetic}
% ======================================================================================== %

The behavior of SA in Transformer-based ASR models has been analyzed in several previous works~\cite{yang20i,usefulness,understanding}.
% The key advantage of SA is its ability to model long-range dependencies by considering all-to-all relationships.
Recently, a study revealed the reason why SA is especially beneficial for ASR~\cite{understanding}.
In a nutshell, SA in lower layers performs phonetic localization that extracts features based on phonological relationships through the whole sequence.
This unique behavior is expected to improve the recognition performance by standardizing the various pronunciation of the same phoneme within the utterance.
The findings on phonetic localization are supported by the phoneme attention relationship (PAR), a tool that visualizes the phonetic behavior of SA by converting frame-to-frame attention to phoneme class-to-class attention~\cite{understanding}.
Specifically, the $(i, j)\text{-th}$ element of PAR indicates how much attention weight (in average) is assigned from $i\text{-th}$ phoneme class to $j\text{-th}$ class.
Please refer to the original paper for more details about PAR~\cite{understanding}.

We investigate PAR of phonetic heads and find that such heads can be further separated into two groups.
Figure~\ref{fig:par} visualizes representative PAR examples.
The first row focuses on the similarity of frames, characterized by symmetric PAR.
For attention heads belonging to this type, the attention weight follows the correlation between phoneme classes of query and key.
On the other hand, the second row focuses on the individual frame, represented as vertical lines in PAR.
In this case, the attention weight highly depends on the phoneme class of key, and therefore might not be sufficiently represented by the query-key dot product.
Note that individual attention heads cannot be clearly separated into two groups; the more accurate interpretation is that one head contains both tendencies with different portions.
The original work on PAR also observed various PAR patterns of phonetic heads~\cite{understanding}, however, did not much investigate this phenomenon.

% ======================================================================================== %
% ======================================================================================== %
\section{Phonetic Self-Attention}\label{sec:method}
% ======================================================================================== %

\begin{table*}[t]
    \centering
    \caption{Phoneme classification accuracy (\%) of different dot product variants evaluated on LibriSpeech dataset.
    M2 is the dot product of the original self-attention, and M5 is the dot product of the proposed phonetic self-attention.}
    \resizebox{0.86\linewidth}{!}{
    \begin{tabular}{c|l|cccc}
        \toprule
        Model   &   Dot-product     & \textit{dev-clean} & \textit{dev-other} & \textit{test-clean} & \textit{test-other} \\
        \midrule
        \midrule
        % M0  & $(xW^Q +b^Q)(xW^K + b^K)^T$                           & 81.91 & 73.41 & 81.83 & 73.66 \\  % yq-yk
        M1  & $(XW_Q)(XW_K)^T$                                      & 81.92 & 73.42 & 81.86 & 73.63 \\  % noq-nok
        % M2  & $(xW^Q +b^Q)(xW^K)^T$                                 & 81.84 & 73.37 & 81.79 & 73.55 \\  % yq-nok
        M2  & $(XW_Q)(XW_K)^T + (XW_K b_Q^T)^T$                   & 81.84 & 73.37 & 81.79 & 73.55 \\  % yq-nok
        % M3  & $(xW^Q)(xW^K + b^K)^T$                                & 81.99 & 73.50 & 81.95 & 73.73 \\  % noq-yk
        % M4  & $(xW^Q)(xW^K)^T + (xw^P)$                    & 81.77 & 73.18 & 81.67 & 73.46 \\  % kp
        M3  & $(XW_Q)(XW_K)^T + (Xc^T)^T$                  & 81.93 & 73.26 & 81.82 & 73.52 \\  % qp
        \midrule
        % \midrule
        % M6  & $(xW^Q)(xW^K)^T + (\phi(xW^S)w^P)$                & 82.15 & 73.60 & 82.07 & 73.86 \\  % qk3-kp
        M4  & $(XW_Q)(XW_K)^T + (\phi(XW_C)c^T)^T$              & 82.40 & 73.89 & 82.25 & 74.20 \\  % qk3-qp
        \midrule
        M5  & $\psi_s((XW_Q)(XW_K)^T) + \psi_c(\phi(XW_C)c^T)^T$    & \textbf{82.66} & \textbf{74.20} & \textbf{82.53} & \textbf{74.48} \\  % qk3-qp-prelu
        \bottomrule
    \end{tabular}
    }
    % \vspace{0.2cm}
    \label{tab:per}
\end{table*}

% ======================================================================================== %
\subsection{Decomposition of Similarity and Content}\label{ssec:decomposition}
% ======================================================================================== %

We distinguish the two important phonetic behaviors by the dependency on other frames.
The first one, \textit{similarity-based} attention, focuses on the similarity between two frames.
The second one, \textit{content-based} attention, focuses more on the content of each frame.
We connect these two different phonetic behaviors to two terms in Eq.~\eqref{eq:new_dp}.
The attention weight $A[i,j]$ is determined by both the similarity between $i,j\text{-th}$ frames and the content of $j\text{-th}$ frame.
These behaviors can be simultaneously performed with vanilla SA, where the original formulation does not clearly separate these two.

We first decompose two behaviors by modifying the dot product in SA.
Specifically, in Eq.~\eqref{eq:new_dp}, we remove the effect of the first term on the second term by replacing the shared weight $W_K$ with a separate parameter $W_C$:
\begin{equation}
    XW_K b_Q^T \rightarrow \phi(XW_C)c^T \label{eq:content},
\end{equation}
where $\phi$ is the Swish~\cite{swish} function and $c \in \mathbb{R}^{1\times d_h}$ is a bias parameter.
We insert the non-linearity function $\phi$ to avoid two parameters ($W_C$ and $c^T$) collapse.

% ======================================================================================== %
\subsection{Non-linear Activation Function}\label{ssec:nonlinear}
% ======================================================================================== %

Next, we apply the PReLU~\cite{prelu} activation function so that the influence of each term can be controlled before adding the two.
PReLU contains a single trainable parameter $\alpha$ that controls the tangent of the negative slope.
\begin{equation}
    \psi_{s,c}(x) =
        \begin{cases}
            x                       & \text{if}\quad x \geq 0 \\
            \alpha_{s,c} \cdot x    & \text{otherwise},
        \end{cases}
\end{equation}
where $\psi_s$ and $\psi_c$ represent PReLU for similarity- and content-based terms, respectively.
We initialize $\alpha$ to 1 for PReLU to behave like an identity function at the beginning of training.

The proposed \textit{phonetic self-attention (phSA)} is the addition of two terms that correspond to two different phonetic behavior:
\begin{equation}
    \psi_s((XW_Q)(XW_K)^T) + \psi_c(\phi(XW_C)c^T)^T.  \label{eq:phsa}
\end{equation}
The first and the second terms represent similarity-based and content-based phonetic attention, respectively.
The proposed phSA is a direct drop-in replacement to the conventional dot product and is easy to implement.

% ======================================================================================== %
\subsection{Additional Design Choices}\label{ssec:design}
% ======================================================================================== %

\subsubsection{Remove Positional Encoding}

The relative positional encoding (RPE) has been widely used for Transformer models for ASR~\cite{transformer-transducer, conformer, rpe-asr, cape}. For example, Conformer~\cite{conformer} exploits the same RPE implementation as Transformer-XL \cite{transformer-xl}.
Although the previous study suggested that RPE may be unnecessary for large size ASR models~\cite{pushing-semi}, RPE helps small to medium size ASR models to better generalize to variable sequence lengths~\cite{pe-jhpark}.
The downside of RPE is the heavy computation cost caused by additional query-position relationship computation and complex tensor operations to match the relative position.
We decide not to use any positional information when using phSA; neither absolute nor relative PE is used.
The design is based on the idea that the phonetic behavior of SA would consider each frame's phonetic characteristics, not necessarily the relative distance between frames.
As a good side effect, the weight parameter for RPE is removed while $W_C$ is added, so the number of parameters in phSA remains almost the same as in SA.
We note that using RPE and phSA together may provide additional gain on performance at the expense of increased resource usage.

\subsubsection{Replace in Lower Layers}

We only replace the vanilla SA with phSA for the lower layers of the model, where phonetic localization is performed~\cite{understanding}.
Because upper layers are known to be responsible for linguistic localization that combines the extracted phonetic information to generate text, we expect phSA may not be useful for those layers.
From the experiments, we show that using phSA for the entire layers actually hurts the performance (see Sec.~\ref{ssec:asr}).
% ======================================================================================== %
% ======================================================================================== %
\section{Experimental Results}\label{sec:experiment}
% ======================================================================================== %

% ======================================================================================== %
\subsection{Setup}\label{ssec:setup}
% ======================================================================================== %

We train our ASR models on the LibriSpeech-960 dataset~\cite{librispeech}.
For both phoneme classification and speech recognition experiments, we employ 80-dimensional log-Mel filterbank features as the input, extracted from a 25ms window with a 10ms stride.
We employ 36 phoneme classes (including `silence') for the phoneme classification as in~\cite{understanding}.
For speech recognition, the subword vocabulary size is set to 128, built by SentencePiece~\cite{sentencepiece} on the training data transcripts.

We choose the Conformer-M~\cite{conformer} as our baseline and train the model with CTC~\cite{ctc} loss.
The baseline Conformer-M consists of 16 Conformer layers with RPE.
We follow the training details from the previous work~\cite{understanding} for ASR.
For the phoneme classification task, we stack 4 Conformer layers with the hidden dimension of 256.
When replacing SA with phSA, we only modify the self-attention block inside the Conformer layer and preserve other blocks such as convolutional and feed-forward blocks.
We set the learning rate to $1.56\text{e-3}$ and weight decay to $1\text{e-4}$ for the phoneme classification.

% ======================================================================================== %
\subsection{Phoneme Classification}\label{ssec:phoneme}
% ======================================================================================== %

To evaluate the phonetic feature extraction performance, we train the models for phoneme classification.
Table~\ref{tab:per} compares the vanilla SA (M2), phSA (M5), and other variants.
M2 is the original dot-product, and M1 is the same version without bias parameter that only focus on similarity-based relationships.
M3 is identical to the M2 but differs in the implementation that the parameter $W_K$ is not shared.
M1, M2, and M3 show almost similar accuracy with less then 0.1\% difference.
In contrast, M4 shows a noticeable gain in phoneme classification accuracy compared to M2 and M3.
The proposed phSA (M5) achieves the highest accuracy among the dot-product variants.
The results verify that our architectural modifications, M2$\rightarrow$M4 (Sec.~\ref{ssec:decomposition}) and M4$\rightarrow$M5 (Sec.~\ref{ssec:nonlinear}), each contributes to better phonetic feature extraction.

\begin{table}[t]
    \centering 
    \caption{Word error rate (\%) of different configurations of phonetic self-attention layers.
    The baseline performance (without phSA) is presented in the first row.
    The best results are in bold, and the second best results are underlined.}
    \resizebox{0.88\columnwidth}{!}{
    \begin{tabular}{cc|cccc}
        \toprule
        \multicolumn{2}{c|}{\#Layers} & \multicolumn{2}{c}{\textit{dev-}} & \multicolumn{2}{c}{\textit{test-}} \\
        
        phSA & SA & \textit{clean} & \textit{other} & \textit{clean} & \textit{other} \\
        \midrule
        \midrule
        0    & 16 & 3.10 & 8.23 & 3.25 & 8.21 \\
        % 1    & 15 & 3.12 & 8.17 & 3.36 & 8.13 \\
        4    & 12 & \textbf{2.87} & 8.11 & \underline{3.19} & \textbf{7.88} \\
        6    & 10 & \underline{3.01} & \textbf{7.77} & \textbf{3.15} & \underline{7.93} \\
        8    & 8  & 3.05 & \underline{8.06} & \underline{3.19} & 8.06 \\
        12   & 4  & 3.08 & 8.36 & 3.30 & 8.30 \\
        % 15   & 1  & 3.31 & 8.87 & 3.53 & 9.00 \\
        16   & 0  & 3.58 & 9.55 & 3.81 & 9.51 \\
        \bottomrule
    \end{tabular}
    }
    \vspace{-0.2cm}
    \label{tab:wer}
\end{table}

% ======================================================================================== %
\subsection{Speech Recognition}\label{ssec:asr}
% ======================================================================================== %

Table~\ref{tab:wer} shows the end-to-end speech recognition performance with the proposed phSA.
Compared to the baseline, replacing the vanilla SA to phSA reduces the word error rate (WER) on every data subset, especially for the challenging LibriSpeech \textit{dev-other} and \textit{test-other} datasets.
We empirically show that adopting phSA only for lower layers (under $8\text{-th}$ layer) achieves the best performance.
This observation is aligned with previous analysis~\cite{understanding} that the lower layers more focus on phonetic information than upper layers.
For example, replacing phSA for lower 6 layers decreases the WER from 8.23\% to 7.77\% (5.6\% relative reduction) and 8.21\% to 7.93\% (3.4\% relative reduction) for \textit{dev-other} and \textit{test-other} datasets, respectively.
In contrast, utilizing phSA for 12 layers shows worse performance than the baseline, and using phSA for the entire (16) layers suffers from significant performance degradation.

% ======================================================================================== %
\subsection{Discussion}\label{ssec:discussion}
% ======================================================================================== %

\subsubsection{Speed and Parameter Size}

Although we add several new computation steps for phSA, we observe that the training and inference time does not change much.
The main reason is that the removal of RPE can compensate for the additional cost of phSA, in both latency and parameter size.
For example, the wall-clock training time of the phSA ($2^\text{nd}$ row in Table~\ref{tab:wer}) is about 5\% faster than the baseline ($1^\text{st}$ row) with almost the same number of parameters.
% Using more phSA layers can further reduce the time cost, but the benefit is marginal.

% \subsubsection{Phoneme Classification from Hidden Representation}

% To verify whether the performance improvement of phSA comes from the enhanced ability to extract the important phonetic information, we conduct a phoneme classification experiment using intermediate hidden representations (features) extracted from the trained end-to-end ASR models. 
% Specifically, we train a softmax classifier consisting of a single fully-connected layer.
% We compare the baseline (zero phSA layer), 4 phSA layer, and 8 phSA layer models presented in $1^\text{st}, 2^\text{nd}$, and $3^\text{rd}$ rows in Table~\ref{tab:wer}.
% Figure~\ref{fig:prelu} shows the phoneme classification accuracy of different layers' features.
% We observe that there is a consistent accuracy gap between the three models, where the model with 8 phSA layers achieves the highest accuracy.
% The result implies that phSA can be more suitable in phonetic feature extraction than the original SA.

\begin{table}[t]
    \centering
    \caption{Effect of similarity-based (S) and content-based (C) attention by removing each component. 
    Entropy (mean $\pm$ std) of phSA attention maps and word error rate (\%) are reported.}
    \resizebox{0.96\columnwidth}{!}{
    \begin{tabular}{cc|cll}
        \toprule
        % \multirow{2}{*}{S} & \multirow{2}{*}{C} & \multirow{2}{*}{Entropy} & \multicolumn{2}{c}{\textit{test-}} \\
        % & & & \textit{clean} & \textit{other} \\
        S & C & Entropy & \textit{dev-other} & \textit{test-other} \\
        \midrule
        \midrule
        \cmark & \cmark & 1.91 $\pm$ 0.12 & 7.77         & 7.93        \\
        \cmark &        & 2.02 $\pm$ 0.14 & 8.16 (+0.39) & 8.54 (+0.61) \\
               & \cmark & 2.39 $\pm$ 0.04 & 9.20 (+1.43) & 9.35 (+1.42) \\
        \bottomrule
    \end{tabular}
    }
    \label{tab:ablation}
    \vspace{-0.2cm}
\end{table}

\subsubsection{Comparison of Similarity and Content}\label{sssec:comparison}

To understand the relative importance between similarity-based and content-based attention, we evaluate the recognition performance without each component.
Table~\ref{tab:ablation} presents the word error rate of the model using only similarity-related or content-related computation.
Specifically, we fix the parameters of the converged model with 6 phSA layers and discard either term of the phSA dot product (Eq.~\eqref{eq:phsa}).
Removing similarity-based attention (bottom row of Table~\ref{tab:ablation}) degrades the performance more than removing content-based one (top row of Table~\ref{tab:ablation}), which implies that the phonetic features extracted from similarity are more important than content-based attention; however, both are indispensable for speech recognition.

In addition, we calculate the average per-head entropy of attention probability for two settings and observe the meaningful difference.
Similarity-based attention probabilities are more concentrated (lower entropy) and content-based attention probabilities are more distributed (higher entropy).
In other words, the similarity-based term emphasizes the difference while the content-based term enhances the uniformness.
We believe that the proposed phSA encourages two terms to be specialized for different attention patterns.

\subsubsection{PReLU Negative Slope}

\begin{figure}[t]
    \centering
    \resizebox{1.0\columnwidth}{!}{
    \includegraphics{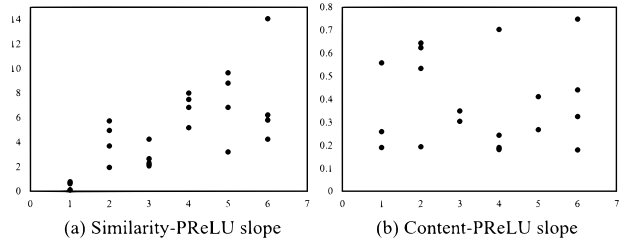}}
    \caption{PReLU negative slopes ($\alpha_s$ and $\alpha_c$) of the phSA after training.
    Four dots in each layer indicates PReLU parameters in four attention heads.
    $x$-axis indicates the layer index.
    Note that the range of $y$-axis is very different, $(0\sim 14)$ for (a) and $(0\sim 0.8)$ for (b).}
    \label{fig:prelu}
    \vspace{0.2cm}
\end{figure}

The range of PReLU negative slopes is very different for similarity-based and content-based terms after training.
Figure~\ref{fig:prelu} shows the negative slopes $\alpha_{s,c}$ of each attention head.
For similarity-based ones, most of $\alpha_s$ parameters are trained to become much larger than 1, implying that the negative correlation values are aggressively ignored and therefore produce a concentrated probability distribution.
On the other hand, $\alpha_c$ parameters for content-based ones are trained to be smaller than 1, decreasing the difference between negative results.
The combination of large $\alpha_s$ and small $\alpha_c$ is connected to different characteristics discussed in Sec.~\ref{sssec:comparison}.

% ======================================================================================== %
% ======================================================================================== %
\section{Related Work}\label{sec:related}
% ======================================================================================== %

Architectural modifications for Transformer-based ASR models have been of great interest.
Many works focus on reducing the heavy computational cost caused by SA~\cite{efficient-conformer, sparse-conformer, simplified-sa, synthesizer}.
For example, Efficient Conformer~\cite{efficient-conformer} proposed grouped SA and downsampling block to shorten the length of the sequence to be processed.
Our work is very distinct from previous works in two points.
First, phSA is designed to enhance the quality of intermediate feature representation, therefore improving the recognition performance.
Second, only a lower part of the model is changed to phSA so that the model utilizes two different types of self-attention mechanisms together.

Pretraining-based approaches have been proven effective in improving the ability to capture useful phonetic information for various downstream tasks.
For example, Wav2Vec2.0~\cite{wav2vec2}, XLSR~\cite{xlsr}, TERA~\cite{tera}, and ACPC~\cite{acpc} presented various self-supervised speech pretraining methods and showed that phonologically meaningful features can be captured while learning the general characteristics of speech.
However, these models use identical Transformer architecture for every layer without considering the different behaviors of each.
Explicit pretraining objectives have also been introduced for learning the useful phonetic features during pretraining.
For example, UniSpeech~\cite{unispeech} and BERTphone~\cite{ling2020bertphone} exploited CTC loss using phoneme sequence as label.
The drawback of the abovementioned studies is that they require an additional pretraining stage before finetuning the model for ASR.

% ======================================================================================== %
% ======================================================================================== %
\section{Conclusion}\label{sec:conclusion}
% ======================================================================================== %

In this paper, we proposed a variant of self-attention (SA), named phonetic self-attention (phSA), to improve the ASR performance.
Especially, we investigated the phonetic behavior of attention heads and distinguished two different attention patterns, similarity-based and content-based attention.
The proposed phSA emphasized the two behaviors by applying simple and effective modifications to the original dot-product in SA.
In addition, the effect of each behavior is controlled by additional trainable parameters.
From the phoneme classification experiments, we showed that phSA is more suitable than the vanilla SA for phonetic feature extraction.
By replacing SA in lower layers with phSA, we improved the speech recognition performance on the end-to-end Transformer-based ASR model.

% ======================================================================================== %
\section{Acknowledgements}
This work was supported by the National Research Foundation of Korea (NRF) grant funded by Korea government (MSIT) (No. 2021R1A2C1013513). This work was also supported in part by Samsung Advanced Institute of Technology, Samsung Electronics Co., Ltd.

\newpage
\bibliographystyle{IEEEtran}
\bibliography{reference}

\end{document}